\documentclass[twoside,leqno,twocolumn]{article}
\usepackage[english]{babel}

\usepackage[letterpaper,top=2cm,bottom=2cm,left=3cm,right=3cm,marginparwidth=1.75cm]{geometry}

\usepackage{amsmath}
\usepackage{graphicx}
\usepackage[colorlinks=true, allcolors=blue]{hyperref}
\usepackage[T1]{fontenc}
\usepackage{amsfonts}
\usepackage{graphicx}
\usepackage{epstopdf}
\usepackage{enumitem}
\usepackage{algorithmic}
\ifpdf
  \DeclareGraphicsExtensions{.eps,.pdf,.png,.jpg}
\else
  \DeclareGraphicsExtensions{.eps}
\fi

\usepackage{amsopn}

\usepackage{authblk}
\usepackage[table]{xcolor}
\definecolor{mygray}{gray}{0.9}
\usepackage{makecell}
\usepackage{booktabs}
\usepackage[utf8]{inputenc}
\usepackage{amsmath,amssymb}
\usepackage{multirow}
\usepackage{array}
\usepackage{titlesec}
\usepackage{adjustbox}
\titlespacing{\section}{0pt}{*1}{*1}
\titlespacing{\subsection}{0pt}{*1}{*1}
\definecolor{lightgreen}{RGB}{220,245,220}
\definecolor{lightred}{RGB}{250,220,220}
\definecolor{lightblue}{RGB}{186,206,230}
\definecolor{title}{HTML}{BFD9BC}
\definecolor{author}{HTML}{B9CDE5}
\definecolor{year}{HTML}{F5CCA0}
\definecolor{venue}{HTML}{CAB1CF}
\definecolor{doi}{HTML}{E7A6A2}
\begin{document}

\title{Where Fake Citations Are Made: Tracing Field-Level Hallucination to Specific Neurons in LLMs}

\author[1]{Yuefei Chen}
\author[1]{Yihao Quan}
\author[1]{Xiaodong Lin}
\author[1]{Ruixiang Tang\thanks{Corresponding author: \texttt{ruixiang.tang@rutgers.edu}}}

\affil[1]{Rutgers University}

\date{}

\maketitle

% Copyright Statement
% When submitting your final paper to a SIAM proceedings, it is requested that you include
% the appropriate copyright in the footer of the paper.  The copyright added should be
% consistent with the copyright selected on the copyright form submitted with the paper.
% Please note that "20XX" should be changed to the year of the meeting.

% Default Copyright Statement
% \fancyfoot[R]{\scriptsize{Copyright \textcopyright\ 2026 by SIAM\\
% Unauthorized reproduction of this article is prohibited}}

%\begin{abstract}
\noindent{\bfseries Abstract}

\noindent LLMs frequently generate fictitious yet convincing citations, often expressing high confidence even when the underlying reference is wrong. We study this failure across 9 models and 108{,}000 generated references, and find that author names fail far more often than other fields across all models and settings. Citation style has no measurable effect, while reasoning-oriented distillation degrades recall. Probes trained on one field transfer at near-chance levels to the others, suggesting that hallucination signals do not generalize across fields. Building on this finding, we apply elastic-net regularization with stability selection to neuron-level CETT values of Qwen2.5-32B-Instruct and identify a sparse set of field-specific hallucination neurons (FH-neurons). Causal intervention further confirms their role: amplifying these neurons increases hallucination, while suppressing them improves performance across fields, with larger gains in some fields. These results suggest a lightweight approach to detecting and mitigating citation hallucination using internal model signals alone.

\section{Introduction}
\label{sec:introduction}
\begin{figure*}[h]
  \centering
  \includegraphics[width=0.9\textwidth]{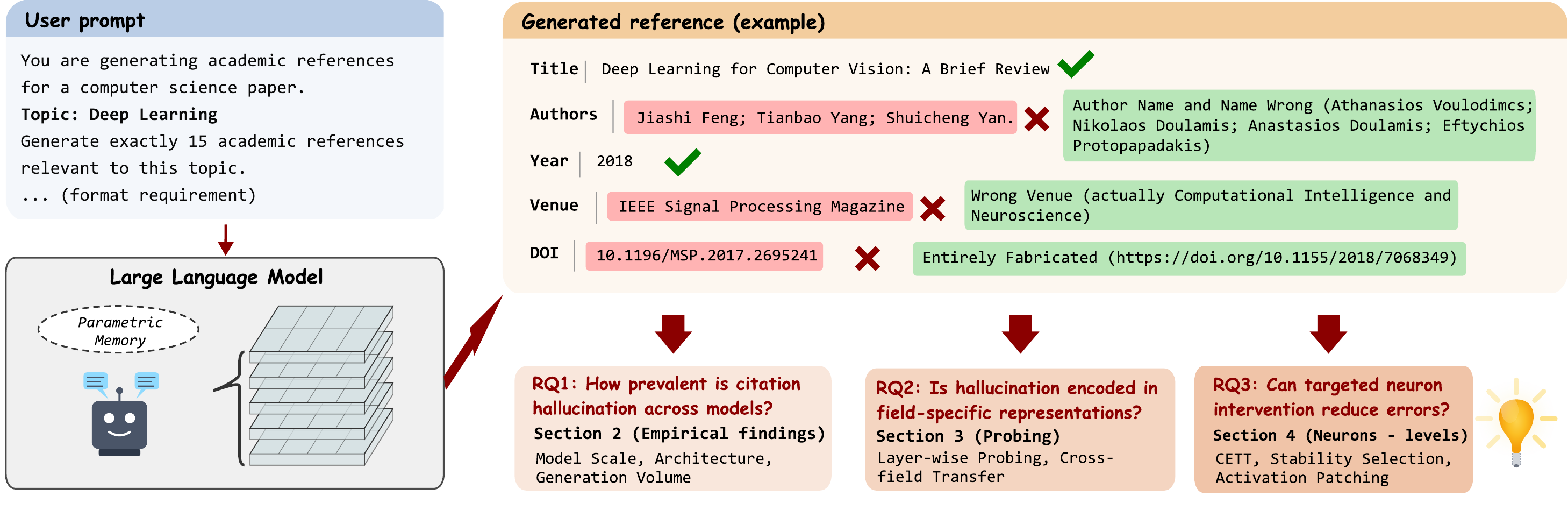}
  \caption{Overview of citation hallucination in LLMs. Given a topic, models generate references with plausible but incorrect metadata. We investigate three research questions: the prevalence of such hallucinations, how they are encoded in model representations, and whether targeted neuron intervention can reduce errors.}
  \label{fig:pipeline}
\end{figure*}
\leavevmode \\
Large language models are increasingly used to draft related work and bibliographies, but when they rely on parametric memory alone they cannot distinguish confident recall from confident fabrication. This leads to a recurring pattern: references that look correct at first glance but contain errors in one or more bibliographic fields. This problem has already appeared in recent work. A recent audit of NeurIPS 2025 accepted papers found over 100 hallucinated citations that went undetected during peer review~\cite{gptzero2026neurips}.
 
Prior work largely focuses on detecting or avoiding hallucination, without explicitly modeling its underlying causes. For example, post-hoc verification pipelines~\cite{min2023factscore,wei2024long} can detect incorrect references, but they require multiple API calls per citation and operate as black-box checks, providing little insight into why the model makes these errors. Retrieval-augmented generation~\cite{lewis2020retrieval} reduces hallucination by grounding outputs in external documents. However, because it relies on external information, it does not address the internal mechanisms that give rise to hallucination. As a result, it remains unclear where citation hallucination arises within the model, and whether it can be detected and corrected using internal model signals.
 
%A bibliographic reference is a composite of structurally distinct fields: author, title, venue, year, and DOI. Each can be wrong for different reasons and may be encoded through different internal mechanisms. Prior interpretability work on hallucination~\cite{azaria2023internal,gao2025hneurons, li2023inference, marks2023geometry,orgad2024llm} has treated it as a monolithic phenomenon, aggregating across all types of factual error. Citation hallucination resists this aggregation. Whether the tools developed for general factoid QA extend to this structured, multi-field setting has not been examined.

This question is especially important for citations because a bibliographic reference is a structured composition of distinct fields, including author, title, venue, year, and DOI. Prior interpretability work has shown that truth-related signals in LLM hidden states can be recovered with simple probes \cite{marks2023geometry, azaria2023internal}, and that activation-level interventions can steer model behavior toward truthfulness~\cite{li2023inference, zou2023representation}. At the neuron level, \cite{gao2025hneurons} identified hallucination-associated neurons and showed that suppressing them reduces errors in factoid QA. Meanwhile, \cite{orgad2024llm} found that truthfulness probes generalize weakly across datasets, suggesting that these internal signals may vary across settings rather than transfer uniformly. Together, these findings highlight the importance of studying hallucination in structured settings such as citations, where a single reference contains multiple co-dependent fields, each of which may fail for different reasons. However, tools developed for general factoid QA cannot be directly applied to this structured, multi-field setting.
 
To address this gap, we conduct three analyses, each organized around a distinct research question (Figure~\ref{fig:pipeline}):

\begin{itemize}
\item \textbf{How prevalent is citation hallucination in different fields?} We build a large-scale dataset by prompting multiple models to generate references across 50 topics and 8 citation styles, and verify each bibliographic field against metadata records from OpenAlex~\cite{priem2022openalex}. Using this dataset, we find a consistent ranking of field-level error rates across models and generation settings, with author names being the most error-prone, followed by venues, titles, and years.
 
\item \textbf{Is hallucination encoded in field-specific representations?} We train linear probes on the hidden states of Qwen2.5-32B-Instruct \cite{bai2023qwen} and find that different fields show different layer-wise patterns. Probes trained on one field perform at near-chance levels on the others, suggesting that citation errors in different fields are represented differently inside the model.
 
\item \textbf{Can targeted neuron intervention reduce field-level errors?} We apply elastic-net regularization with stability selection to neuron-level CETT contributions and identify a small set of field-specific hallucination neurons (FH-neurons). When we amplify these neurons, hallucination increases. When we suppress them, accuracy improves, and this improvement does not appear under random ablation. This provides causal evidence that specific neurons contribute to field-level citation errors.
\end{itemize}
 
\noindent Together, these findings show that citation hallucination is detectable from internal model signals, follows different patterns across bibliographic fields, and can be partly reduced through targeted neuron suppression without external retrieval.
\begin{table*}[!htbp]
\centering
\scalebox{0.53}{
\begin{tabular}{p{5.5cm}|>{\centering\arraybackslash}p{1.2cm}>{\centering\arraybackslash}p{ 0.8cm}>{\centering\arraybackslash}p{ 0.7cm}>{\centering\arraybackslash}p{1cm}>{\centering\arraybackslash}p{ 0.6cm}>{\centering\arraybackslash}p{ 0.75cm}|>{\centering\arraybackslash}p{1.2cm}>{\centering\arraybackslash}p{ 0.7cm}>{\centering\arraybackslash}p{ 0.7cm}>{\centering\arraybackslash}p{0.75cm}>{\centering\arraybackslash}p{ 0.7cm}>{\centering\arraybackslash}p{ 0.7cm}|>{\centering\arraybackslash}p{1.2cm}>{\centering\arraybackslash}p{ 0.7cm}>{\centering\arraybackslash}p{ 0.7cm}>{\centering\arraybackslash}p{0.7cm}>{\centering\arraybackslash}p{ 0.7cm}>{\centering\arraybackslash}p{ 0.75cm}}
\toprule
\rowcolor{mygray} \multicolumn{1}{c}{\textbf{N (Refs/Prompt)}} & \multicolumn{6}{c}{\textbf{N=5}} & \multicolumn{6}{c}{\textbf{N=10}} & \multicolumn{6}{c}{\textbf{N=15}} \\ \hline
\rowcolor{mygray} \textbf{Models} & \textbf{Author} & \textbf{Title} & \textbf{Year} & \textbf{Venue} & \textbf{DOI} & \textbf{Total} & \textbf{Author} & \textbf{Title} & \textbf{Year} & \textbf{Venue} & \textbf{DOI} & \textbf{Total} & \textbf{Author} & \textbf{Title} & \textbf{Year} & \textbf{Venue} & \textbf{DOI} & \textbf{Total} \\ \hline

%\multicolumn{19}{l}{\textit{Qwen}} \\ 
\hline
Qwen2.5-14B-Instruct & 9.3 & 39.1 & 32.8 & 25.4 & 22.7 & 4.3 & 7.7 & 31.6 & 28.9 & 23.7 & 21.0 & 3.2 & 7.1 & 29.1 & 27.4 & 23.0 & 20.0 & 2.9 \\
Qwen2.5-32B-Instruct & 16.0 & 45.8 & 43.0 & 39.8 & 54.9 & 10.5 & 13.6 & 43.4 & 41.6 & 39.3 & 53.8 & 8.4 & 13.1 & 40.2 & 40.6 & 38.6 & 53.5 & 8.0 \\
Qwen3-30B-A3B-Base & 19.2 & 57.1 & 51.9 & 38.9 & 27.7 & 5.0 & 13.3 & 42.1 & 41.3 & 31.7 & 20.7 & 3.1 & 10.3 & 32.5 & 34.0 & 26.6 & 17.7 & 2.3 \\
Qwen3-30B-A3B-Instruct & 15.5 & 55.2 & 47.3 & 33.0 & 33.3 & 6.8 & 11.5 & 43.1 & 39.1 & 28.6 & 25.4 & 4.5 & 9.1 & 34.4 & 33.8 & 25.9 & 20.2 & 3.1 \\
\midrule
%\multicolumn{19}{l}{\textit{Others}} \\ 
%\hline
Moonlight-16B-A3B-Instruct & 9.6 & 44.8 & 35.2 & 26.2 & 38.6 & 4.3 & 7.1 & 37.0 & 29.3 & 21.4 & 32.7 & 2.5 & 6.2 & 31.2 & 27.8 & 20.4 & 29.8 & 2.0 \\
Mistral-Small-24B-Instruct-2501 & 15.4 & 52.7 & 45.5 & 33.2 & 32.6 & 5.0 & 10.2 & 42.5 & 38.5 & 27.5 & 25.0 & 2.9 & 8.2 & 34.1 & 33.6 & 25.5 & 20.4 & 1.9 \\
\midrule
DeepSeek-R1-Distill-Qwen3-8B & 20.4 & 39.8 & 44.1 & 29.0 & 23.7 & 3.2 & 12.0 & 28.3 & 31.0 & 21.2 & 25.0 & 2.2 & 10.2 & 22.5 & 25.5 & 17.5 & 25.5 & 2.2 \\
DeepSeek-R1-Distill-Qwen2.5-14B & 7.7 & 22.0 & 26.4 & 19.8 & 13.2 & 1.1 & 3.9 & 12.7 & 21.5 & 13.8 & 8.3 & 0.6 & 3.4 & 10.9 & 21.5 & 12.5 & 6.4 & 0.4 \\
DeepSeek-R1-Distill-Qwen2.5-32B & 19.8 & 45.5 & 47.5 & 38.6 & 23.8 & 4.0 & 14.4 & 39.6 & 38.5 & 31.0 & 21.4 & 3.2 & 13.3 & 35.8 & 36.5 & 29.2 & 18.8 & 3.0 \\
\bottomrule
\end{tabular}
}
\caption{Field-level verification accuracy across different models and number of references per prompt (N).}
\label{tab:hallucination_rates}
\end{table*}
\section{Empirical Analysis of Citation Hallucination}
% % \leavevmode \\
\subsection{Data Collection}
\label{sec:data}

We prompt a set of large language models to generate academic citations from parametric memory alone for 50 computer science research topics spanning machine learning, NLP, systems, security, and theory (full list in Appendix~\ref{app:topics}). For each topic, the models generate $N \in \{5, 10, 15\}$ references under 8 citation styles: APA, MLA, Chicago, Harvard, Vancouver, IEEE, ACM, and AMA. This design allows us to examine how citation style and generation volume affect hallucination rates. For each model, this procedure yields approximately 12{,}000 generated references across 50 topics, 3 generation volumes, and 8 citation styles. Outputs are constrained by a JSON schema with five structured fields per reference: \texttt{title}, \texttt{authors}, \texttt{venue}, \texttt{year}, and \texttt{doi}. Each reference is treated as an independent example and is subsequently verified and labeled.

\begin{figure}[h]
  \centering
  \includegraphics[width=0.49\textwidth]{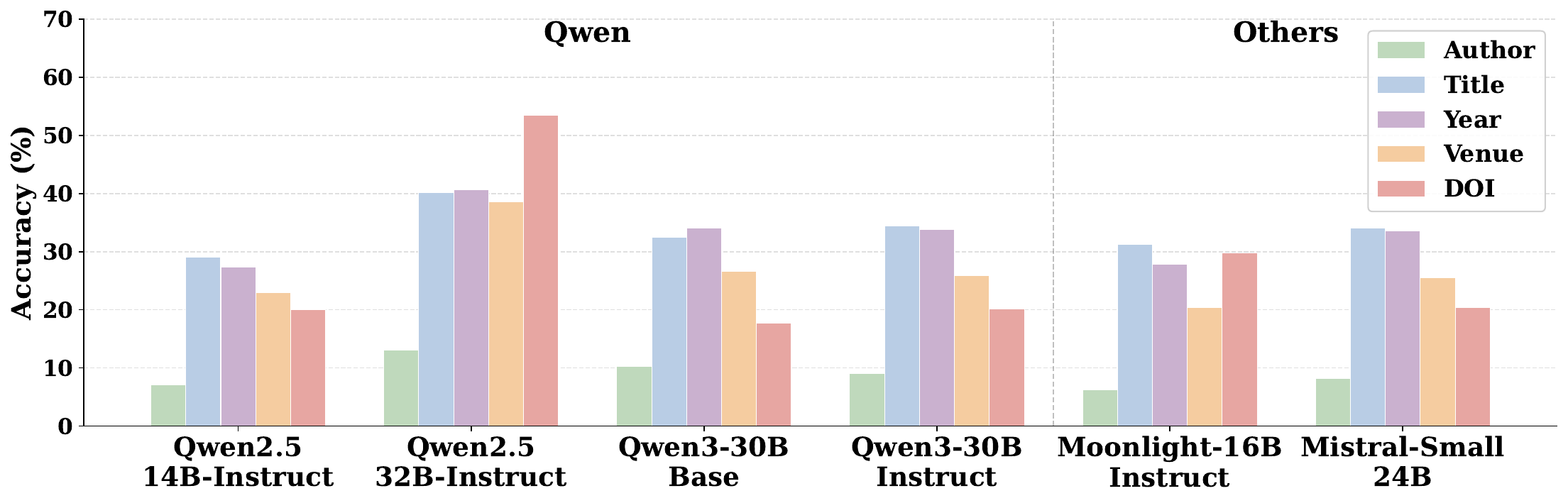}
  \caption{Per-field citation accuracy at $N=15$ across all models.
Author accuracy is consistently the lowest across all models, while
DOI accuracy is notably higher for Qwen2.5-32B-Instruct than for
all other models.}
  \label{fig:field_acc}
\end{figure}
\subsection{Verification Pipeline}
\label{sec:verification}

We first verify each generated reference against OpenAlex~\cite{priem2022openalex} through its public REST API. We use a two-stage lookup procedure. If a DOI is present, we query OpenAlex directly using the normalized DOI. Otherwise, we retrieve the top 10 candidates through title search and select the best match based on title similarity, first-author overlap, and year proximity. This step assigns each reference a binary label for each field, together with a global verdict of \textsc{Supported}, \textsc{Partial}, or \textsc{Unsupported}.

Some cases remain ambiguous after this first stage. A reference may appear \textsc{Partial} because the model cites an arXiv preprint while OpenAlex records the published version, or vice versa. A reference may also appear \textsc{Unsupported} because the work has not yet been indexed in OpenAlex rather than because it is fabricated. To resolve these cases, we introduce a second verification stage using GPT-5.4-mini~\cite{OpenAI2026models} with web search access. For each \textsc{Partial} or \textsc{Unsupported} reference, the verifier retrieves live search results and compares them with the OpenAlex result. When the web-grounded evidence clearly supports a different verdict, we use the web-based result as the final label. Because this verifier relies on retrieval rather than parametric memory, it serves as a grounded judge rather than reproducing the same error mode under evaluation. GPT-5.4-mini is also not among the citation-generation models in our benchmark, which avoids direct model overlap in the verification pipeline. To validate the full pipeline, two expert annotators independently reviewed a random sample of 200 labels. Their judgments agreed with the automated verdicts in 93\% of cases, suggesting that the two-stage procedure reliably distinguishes genuine hallucinations from database coverage artifacts.

\subsection{Model and Generation Factors}
\label{sec:variables}
With this verification pipeline in place, we then study how model design and generation settings affect citation hallucination. We vary four factors. First, we compare Qwen2.5-14B-Instruct and Qwen2.5-32B-Instruct to isolate the effect of scale within the same model family. Second, we compare Qwen2.5-32B-Instruct with Qwen3-30B-Instruct~\cite{yang2025qwen3} to examine differences across model versions. Third, we include Qwen3-30B-Base together with its instruct counterpart to assess the effect of alignment training. Finally, we vary the number of references requested per prompt across $N \in \{5, 10, 15\}$ to test whether hallucination increases as the model is asked to recall more entries at once.
% % A broader comparison with additional
% % open-source models is provided in the appendix.

\subsection{Results: Hallucination Rates and Patterns}
\label{sec:results}

Table~\ref{tab:hallucination_rates} reports per-field accuracy
across all models and generation volumes. Even the strongest models
produce errors on a substantial fraction of references, with author
and DOI fields consistently showing the lowest correct-generation rates.
We analyze these results along three dimensions below.

\textbf{Field-level Analysis}
Table~\ref{tab:hallucination_rates} and Figure~\ref{fig:field_acc} reveal a consistent hierarchy of field difficulty. Across all models and generation volumes, the author field is by far the most error-prone, with correct generation rates below 14\% at $N{=}15$. Title and year achieve broadly similar accuracy in the 27--41\% range, with no consistent ordering between them. Venue lags slightly behind in most configurations. DOI accuracy is the most model-dependent: Qwen2.5-32B-Instruct reaches 53.5\%, suggesting it has internalized a disproportionately large number of exact identifiers, whereas all other models fall in the 17--30\% range.

Crucially, the ordering author $<$ venue $<$ title $\approx$ year holds across every model and generation volume in the table, suggesting that field difficulty reflects a structural property of how these models encode bibliographic knowledge rather than an artifact of any particular configuration. This hierarchy is compounded by a positional effect: accuracy is highest for the first two citations in a prompt and degrades sharply thereafter, indicating that models exhaust their most reliable parametric memory early regardless of field (Appendix~\ref{app:Output_Volume_Analysis}).

\begin{figure}[t]
  \centering
  \includegraphics[width=0.45\textwidth]{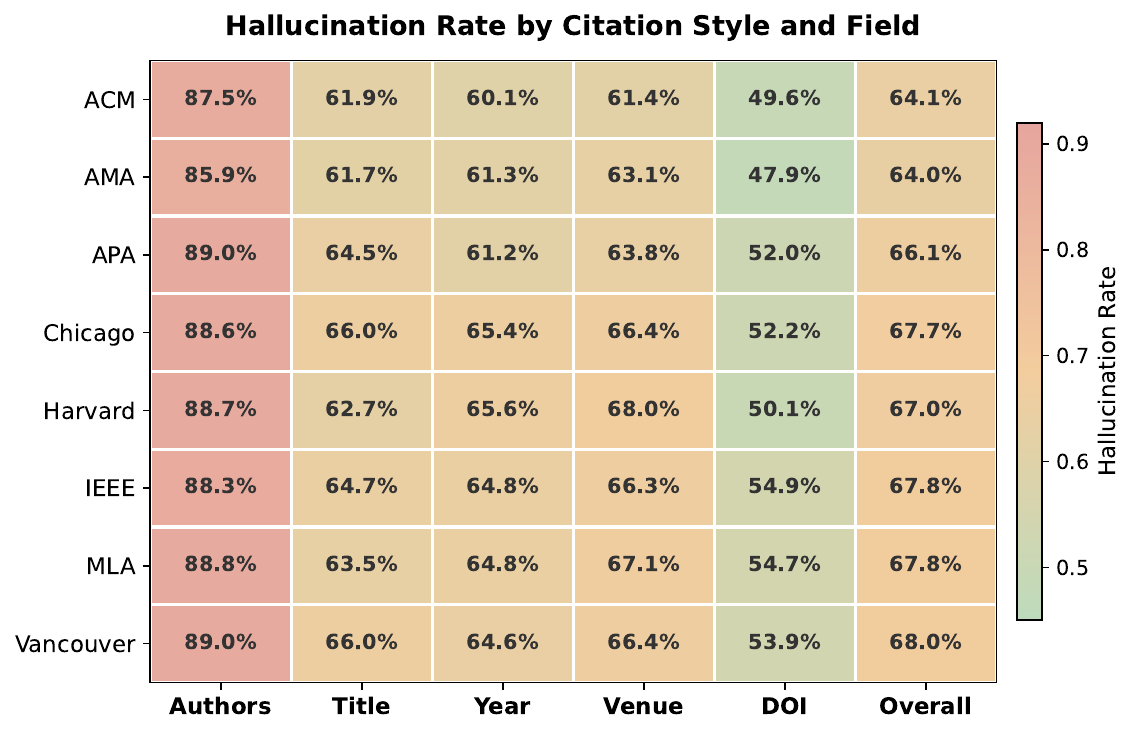}
  \caption{Hallucination rate by citation style and field. Each axis
  corresponds to a citation style; lines represent individual
  bibliographic fields. Author hallucination rates (annotated) are
  consistently the highest across all styles, while style itself has
  negligible effect on any field.}
  \label{fig:radar}
\end{figure}

\textbf{Citation Style-level Analysis}
Figure~\ref{fig:radar} presents hallucination rates across eight citation styles and five bibliographic fields. The author field consistently exhibits the highest hallucination rate across all styles, ranging from 0.86 (AMA) to 0.89 (APA, Chicago, Harvard, Vancouver, and MLA), while DOI shows the lowest and most variable rates. The remaining fields cluster closely in the mid-range, indicating that models achieve similar reliability on these fields regardless of format.

Citation style itself has negligible influence on hallucination behavior.
The maximum observed difference in hallucination rate between any two
styles is less than 0.04 across all fields, and Kruskal-Wallis tests
confirm that none of these differences reach statistical significance
(all $p > 0.05$; see Appendix~\ref{app:style_test} for full results).
This indicates that hallucination is governed by the nature of the
bibliographic field rather than the formatting conventions of the
citation style.

\begin{figure}[t]
  \centering
  \includegraphics[width=0.47\textwidth]{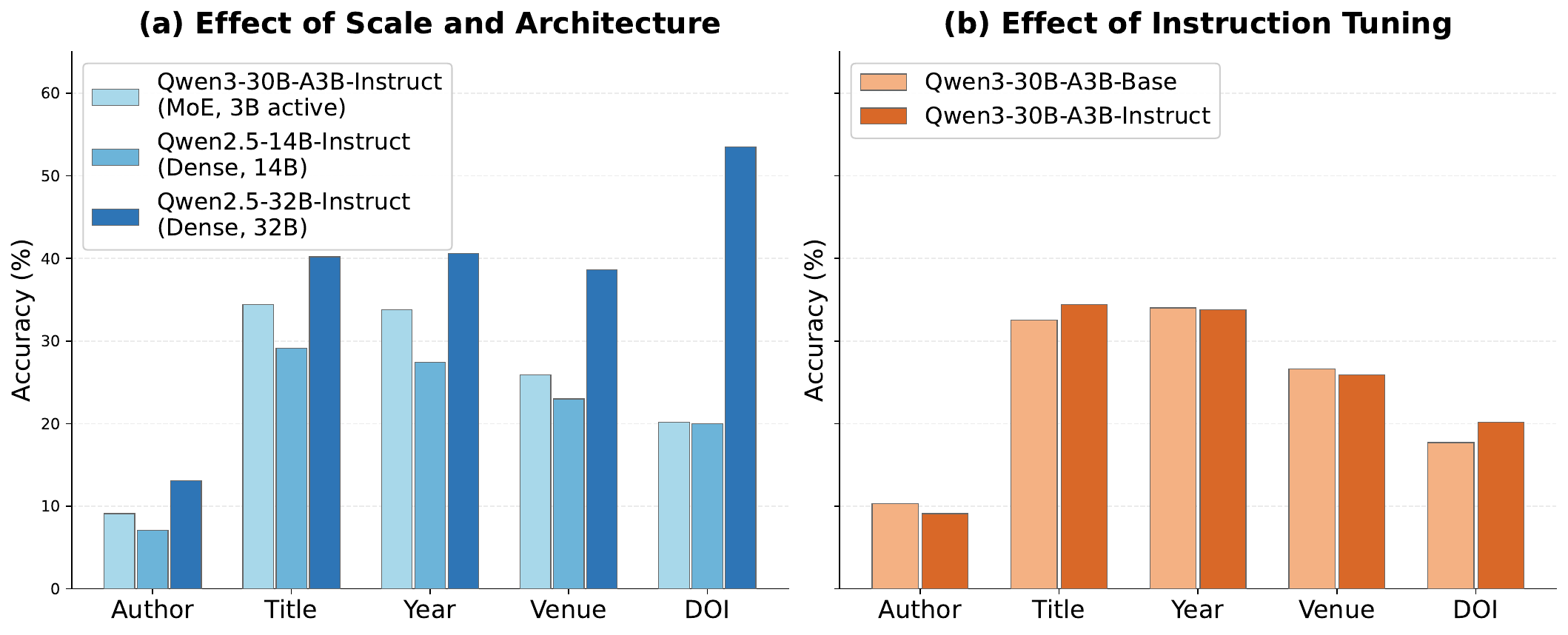}
  \caption{Per-field accuracy under model-level comparisons at $N{=}15$. (a)~Qwen2.5-32B-Instruct (dense, 32B) leads on all fields. (b)~Instruction tuning produces negligible differences across all fields.}
  \label{fig:model_compare}
\end{figure}

\textbf{Model-level Analysis.} 
%\paragraph{Effect of scale and architecture.} 
As shown in Figure~\ref{fig:model_compare}(a), we compare three models spanning a range of effective parameter counts: Qwen3-30B-A3B-Instruct (MoE, 3B active parameters), Qwen2.5-14B-Instruct (dense, 14B), and Qwen2.5-32B-Instruct (dense, 32B). Despite activating only 3B parameters per token, the MoE model matches or slightly exceeds the 14B dense model on title, year, and venue, suggesting that its larger total parameter pool partially compensates for the sparse activation. However, Qwen2.5-32B-Instruct substantially outperforms both, with the largest gains on DOI and venue. Authors accuracy remains below 14\% for all three models, indicating that neither dense scaling nor MoE capacity resolves the most difficult recall task. Moonlight-16B-A3B-Instruct and Mistral-Small-24B-Instruct-2501 follow the same field-difficulty hierarchy, with author accuracy below 10\% and DOI showing the largest between-model spread.

%\paragraph{Effect of reasoning distillation.}
Additionally, three DeepSeek-R1-Distill variants \cite{guo2025deepseek} reveal that reasoning-oriented distillation degrades citation recall. DeepSeek-R1-Distill-Qwen2.5-14B achieves only 0.4\% total accuracy at $N{=}15$, well below its non-distilled counterpart Qwen2.5-14B-Instruct, and DeepSeek-R1-Distill-Qwen2.5-32B trails Qwen2.5-32B-Instruct on every field, with the largest gap on DOI. This pattern suggests that chain-of-thought distillation prioritizes reasoning structure at the expense of factual memorization.

\begin{figure*}[h]
  \centering
  \includegraphics[width=0.8\textwidth]{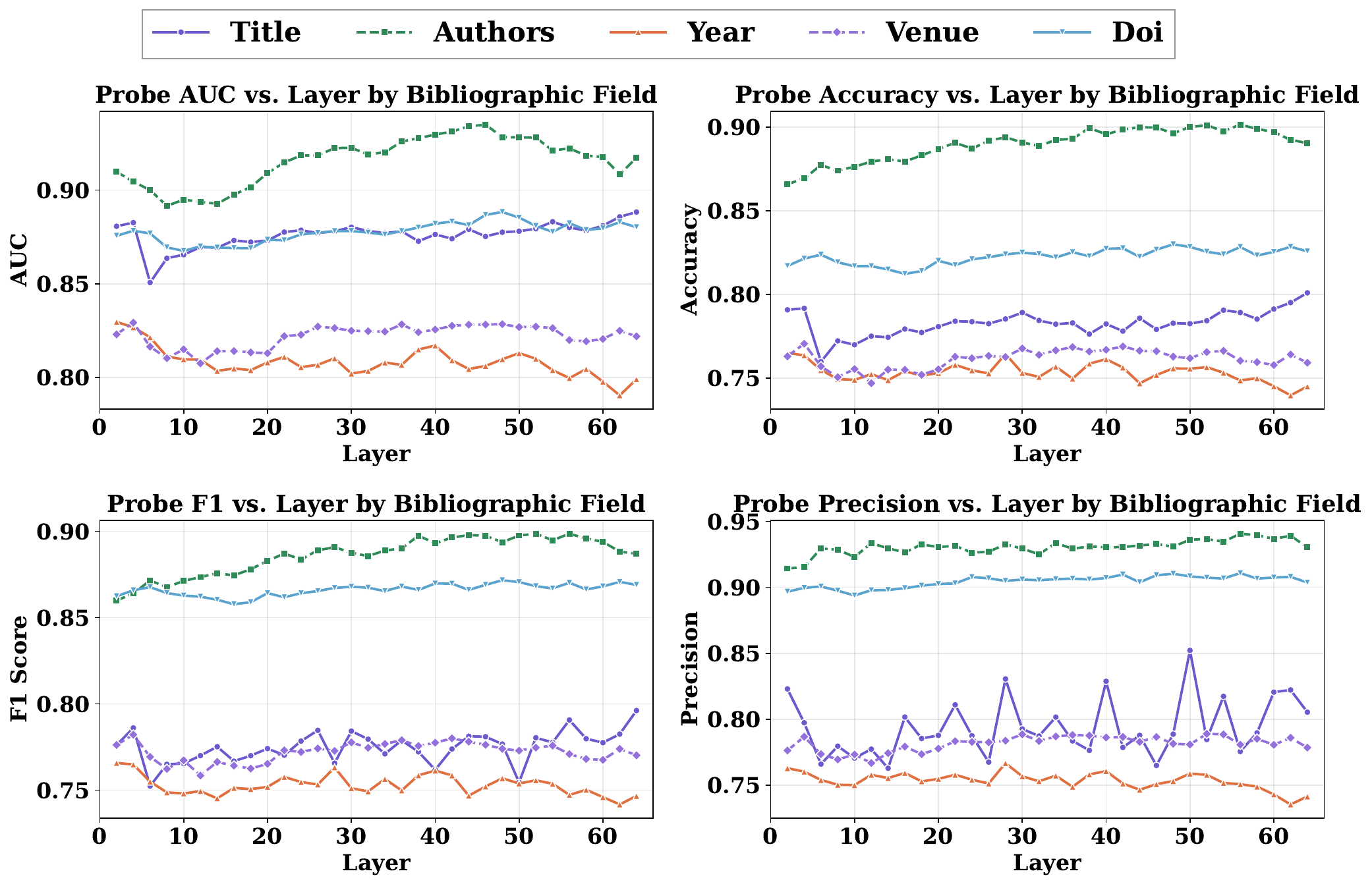}
  \caption{Probe AUC across transformer layers for each bibliographic field in Qwen2.5-32B-Instruct.}
  \label{fig:layer_auc}
\end{figure*}

%% ===================================================================
\section{Probing for Field-Level Hallucination}
%% ===================================================================
\leavevmode \\
The preceding analysis establishes that citation hallucination is pervasive across model families and generation settings, with certain fields like author lists consistently more error-prone than others. These findings raise a deeper question: \textit{do hallucinated citations arise only at the decoding surface, or has the model already committed to an erroneous output within its internal representations?} If the latter is the case, it should be possible to \emph{read off} hallucination from hidden states before any output token is produced.

We focus probing and neuron-localization analyses on Qwen2.5-32B-Instruct because it is the strongest citation generator in Table~\ref{tab:hallucination_rates} and provides open-weight access for hidden-state and neuron-level analysis. To examine whether the cross-field transfer pattern extends beyond this model family, we additionally replicate the transfer heatmap on Mistral-Small-24B-Instruct-2501 \cite{mistral_small_3_2025} in Appendix~\ref{app:Mistral-Small-24B}. We then train token-level linear probes on the hidden states of Qwen2.5-32B-Instruct and pose three progressively sharper questions:

\begin{itemize}
    \item Is citation hallucination decodable from internal representations at all?
    \item How does the hallucination signal evolve across transformer layers, and do different bibliographic fields exhibit distinct layer-wise profiles?
    \item Can a probe trained on one field generalize to another, in other words does each field rely on the same internal mechanism?
\end{itemize}
To answer these questions, we train linear probes on hidden states extracted from field-specific token spans to test whether hallucination signals are linearly decodable across layers. The procedure consists of three steps: (i) serializing each reference with explicit field markers, (ii) extracting hidden states for tokens within each field span, and (iii) training field-specific probes to detect hallucination signals. This design enables layer-wise and cross-field analysis, as detailed below:
\begin{figure*}[htbp]
  \centering
  \includegraphics[width=0.8\textwidth]{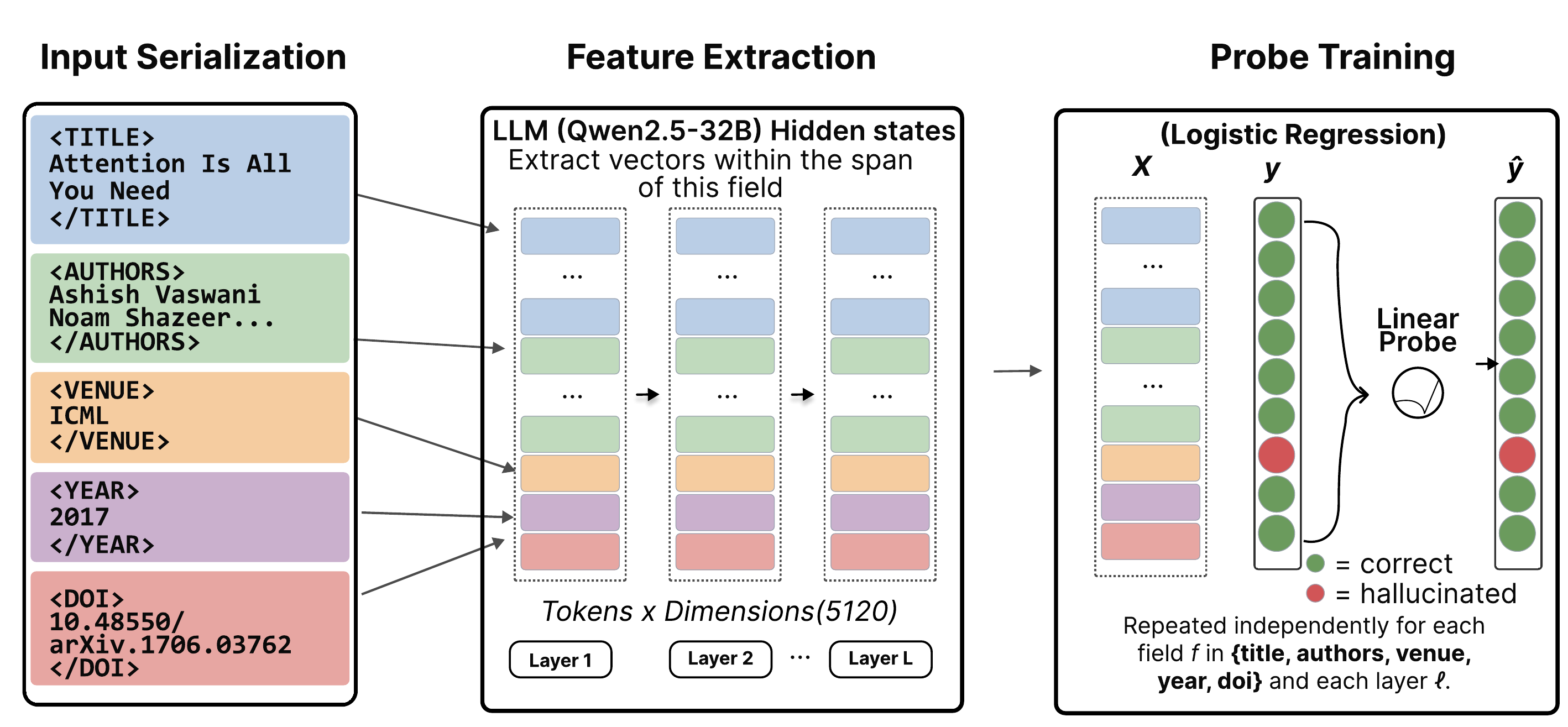}
  \caption{Probing pipeline for citation hallucination detection.}
  \label{fig:probepipeline}
\end{figure*}
\subsection{Input Serialization}
\label{sec:serialization}
 
Each generated reference is serialized into a plain-text sequence in which every bibliographic field is enclosed by XML-style tags (see Appendix~\ref{app:serialization} for an example). During serialization, the character-level start and end offsets of each tagged block are recorded and later mapped to token positions via the tokenizer's offset mapping. This yields precise token spans for every field, enabling the field-specific hidden-state extraction described next.

\subsection{Feature Extraction}
\label{sec:features} 
Given a serialized reference, the text is tokenized and passed through the model with all hidden states retained. For a target field $f$ with token span $[t_s, t_e)$, we collect the hidden-state vectors $\{\mathbf{h}_\ell^{(t)}\}_{t=t_s}^{t_e-1}$ at each layer $\ell$. Each vector serves as an individual training instance, paired with the binary hallucination label of field $f$. Features are strictly field-specific: probing for title hallucination uses only hidden states over the \textsc{title} span, probing for venue uses only the \textsc{venue} span, and so on.

\subsection{Probe Training and Evaluation}
We train $\ell_2$-regularized logistic regression probes (L-BFGS, balanced class weights) independently for each field and layer. A linear model is chosen deliberately: if hallucination is detectable by a linear probe, the signal must be linearly accessible in hidden-state space, a stronger claim than nonlinear detectability. To prevent topic leakage, all references from the same topic are assigned to either training or test, never both, with a 1:1 class balance and an 80\%/20\% topic-level split. To test whether fields share a common hallucination signal, we evaluate each field's probe on every other field's test data; near-chance cross-field AUC would confirm field-specific encoding (details in Appendix~\ref{app:probe_details}). The pipeline is shown in Figure~\ref{fig:probepipeline}.

\subsection{Layer-wise and Cross-field Probing Analysis}
% \textbf{Layer-wise Hallucination Signal}
\label{sec:layer_signal}
% Figure~\ref{fig:layer_auc} plots probe AUC across all transformer layers
% for each citation field. We compare performance across layers to
% identify where in the forward pass hallucination signals are strongest,
% and whether the optimal layer differs across fields. A mid-layer peak
% would suggest that intermediate representations consolidate factual
% knowledge before it is projected to the output vocabulary, while
% field-specific peak layers would indicate that different bibliographic
% fields are retrieved and consolidated at different stages of the forward
% pass.
% \textcolor{red}{RX: This subsection describes the \emph{methodology} and the \emph{hypothesis space} but never reports what the probe actually found. Add concrete numbers: (1) the peak AUC per field, (2) the peak layer per field (does Author peak earlier than DOI?), (3) whether peak layers differ significantly across fields, and (4) a one-sentence interpretation. Without these, Section~3.4.1 reads like a methods paragraph and reviewers will flag it as an incomplete results section.}

Figure~\ref{fig:layer_auc} plots probe AUC across all 64 
transformer layers for each citation field. All five fields 
share a common early-layer dip: AUC starts relatively high 
at layers~2--4, drops through layers~6--10, then diverges 
sharply by field. Authors exhibits the strongest recovery, 
climbing from a minimum of 0.892 at layer~8 to a peak of 
0.935 at layer~46 before declining in the final layers, a 
swing of over four points that indicates substantial 
mid-layer consolidation of author knowledge. Title follows 
a similar but more gradual recovery, reaching its peak of 
0.888 only at layer~64, the sole field whose signal 
strengthens monotonically through the final layers. DOI 
and Venue are comparatively stable across depth, with DOI fluctuating narrowly around 
0.878 and Venue around 0.824. Year is the only field whose 
signal degrades monotonically after the initial layers, 
declining from 0.830 at layer~2 to 0.790 by layer~62.

Spearman rank correlations \cite{spearman1961proof} confirm that these trajectories 
are statistically distinct: Authors, Title, and DOI show 
significant positive trends with depth ($\rho \geq +0.55$, 
all $p{<}0.001$), while Year trends negatively 
($\rho{=}-0.52$, $p{<}0.01$). Fisher $z$-tests show that 
Year's trend differs significantly from every other field 
(all $p{<}0.001$), and a permutation test rejects the null 
that all fields share the same layer--AUC profile 
($p{<}0.0001$). Bootstrap 95\% confidence intervals for
the peak layer are non-overlapping across four of five
fields, spanning from layer~2 (Year) to layer~64 (Title),
while Venue's wide interval covering from layer 4 to 48 reflects its
flat profile (full statistical details in
Appendix~\ref{app:layer_stats}).

\begin{figure}[h]
  \centering
  \includegraphics[width=0.4\textwidth]{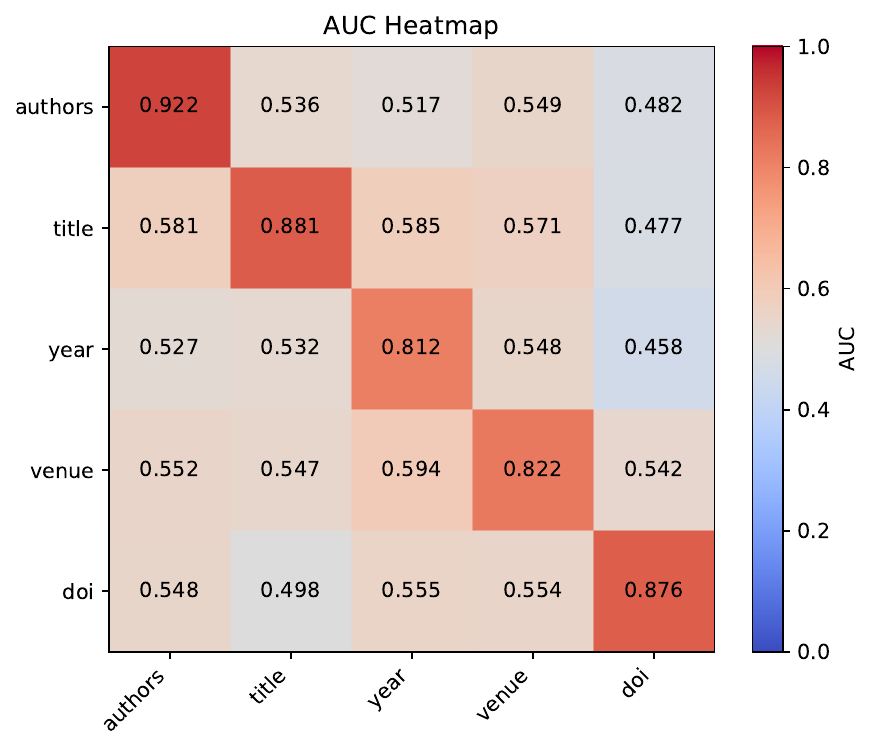}
  \caption{Cross-field AUC heatmap for Qwen2.5-32B-Instruct. Diagonal entries (in-field) range from 0.812 to 0.922, while off-diagonal entries remain near chance (0.46--0.59), indicating that hallucination signals do not transfer across bibliographic fields.}
  \label{fig:heatmap}
\end{figure}

% \textbf{Cross-field Transfer}
We next test whether the hallucination signal is shared across fields or encoded independently. Figure~\ref{fig:heatmap} presents a $5{\times}5$ cross-field AUC heatmap, where entry $(i,j)$ reports the AUC of a probe trained on field $i$ and evaluated on the test set of field $j$. Diagonal entries (in-field performance) range from 0.812 to 0.922, confirming that hallucination is reliably decodable within each field. Off-diagonal entries, however, remain near chance (0.46--0.59), indicating that a probe trained on one field's hallucination signal carries almost no predictive power for any other field. This gap demonstrates that each field's hallucination is encoded in a structurally distinct subspace of the model's representation, rather than a shared signal that generalizes across bibliographic fields.
\label{sec:probe_results}

\section{Field-Specific Neuron Localization}
\label{sec:neuron-localization}
\leavevmode \\
The probing results establish that hallucination is 
encoded in a field-specific, linearly accessible manner, 
but do not identify which individual neurons carry the 
signal. Pinpointing these neurons is a prerequisite for 
causal intervention. We develop a two-stage pipeline 
that isolates a sparse, stable set of per-field 
hallucination neurons from the full CETT feature space 
and tests their causal role via activation patching.We term the neurons identified through this procedure \textit{field-specific hallucination neurons} (FH-neurons).

For each FFN layer $\ell$ and intermediate neuron $n$, CETT is defined as
\begin{equation}
\label{eq:cett}
\text{CETT}_{\ell,n} \;=\; \frac{|a_n| \cdot \|W_{\text{down}}^{(\ell)}[:,n]\|_2}{\|y^{(\ell)}\|_2},
\end{equation}
where $a_n$ is the neuron's pre-projection activation, $W_{\text{down}}^{(\ell)}$ is the down-projection weight matrix, and $y^{(\ell)}$ is the FFN output vector. We compute CETT for every valid token across all 64 layers via forward hooks on each layer's down-projection, yielding a 1.77M-dimensional feature vector per token ($64$ layers $\times$ $27{,}648$ neurons). To obtain a single feature representation per reference, we average the per-token CETT values across all tokens within the target field's span, yielding one vector per reference rather than per token. Training and test sets are split at the topic level to prevent information leakage, and within each partition the two classes are balanced by downsampling.

\subsection{Selection Pipeline}
\label{sec:selection-pipeline}

The CETT feature space is both high-dimensional and highly correlated across neighboring neurons, making direct feature selection unstable. We therefore decompose FH-neuron identification into two stages: sparse candidate selection via elastic-net regression, followed by stability filtering across bootstrap resamples to retain only neurons whose selection is robust to the particular data split.

\textbf{Stage 1: Sparse selection via elastic-net regression.}
We train a logistic regression with an elastic-net penalty over the full CETT feature vector. The regularization combines an $\ell_1$ term with a small $\ell_2$ term:
\begin{equation}
\label{eq:elasticnet}
\mathcal{L} \;=\; \text{BCE}(\hat{y}, y) \;+\; \alpha\,r\,\lVert w \rVert_1 \;+\; \frac{\alpha\,(1-r)}{2}\,\lVert w \rVert_2^2,
\end{equation}
where $r = 0.8$ is the $\ell_1$ ratio. The $\ell_1$ component is optimized via SGD with a proximal soft-thresholding step after each gradient update, while the $\ell_2$ component enters through standard gradient descent. The small $\ell_2$ regularization improves stability when CETT features are correlated across neighboring neurons, encouraging grouped selection without sacrificing the sparsity induced by $\ell_1$. The surviving non-zero entries nominate candidate (layer, neuron) pairs whose CETT values are most predictive of hallucination.

The regularization strength $\alpha$ is selected via grid search, scoring each candidate by a composite criterion that balances validation AUC against sparsity. We adopt AUC rather than accuracy as the detection metric because our goal is to identify neurons that \emph{reliably discriminate} hallucinated from correct fields; AUC captures this ranking quality and is less sensitive to the choice of decision threshold. The score penalizes the proportion of non-zero neurons to favor parsimonious selections while preserving discriminative power. We train one such model per field, yielding a field-specific set of candidate FH-neurons.

\textbf{Stage 2: Stability selection with permutation contro}
The candidates identified by a single elastic-net fit may be sensitive to the particular data split. To guard against this, we apply stability selection: we repeat the elastic-net regression across 20 bootstrap resamples drawn with a 50\% subsample ratio, and record how frequently each neuron is selected across resamples. Only neurons whose selection frequency exceeds 60\% are retained as stable FH-neurons. Among the stable set, we further restrict to neurons with \emph{positive} regression weights, since a positive coefficient indicates that higher CETT activation of that neuron is associated with increased hallucination probability; these are the \textit{pro-hallucination} neurons targeted by our intervention.

After stability selection, this procedure retains 224, 78, 129, 51, and 30 positive-weight FH-neurons for Title, Authors, Year, Venue, and DOI respectively, out of 1{,}769{,}472 candidates per field (at most 0.013\% of the feature space). These neurons are not uniformly distributed across layers. We divide the 64 layers into three equal bands: early (layers 0--21), middle (layers 22--42), and late (layers 43--63). Authors FH-neurons concentrate in the middle band (60.3\%), DOI neurons cluster in the early band (66.7\%), while Title and Year neurons skew toward the late band (41.5\% and 46.5\%). Venue neurons are the most dispersed, spread across 30 layers with no layer exceeding 4 neurons. The full per-layer distribution is provided in Table~\ref{tab:fh_neuron_dist} 
(Appendix~\ref{app:fh_neurons}).
% Neurons with negative weights (anti-hallucination) are excluded from intervention.

Following the stability selection framework of Meinshausen and Bühlmann  \cite{meinshausen2010stability}, the expected number of false discoveries is bounded by
\begin{equation}
\mathbb{E}[V] \;\leq\; \frac{q^2}{(2\pi_{\mathrm{thr}} - 1) \cdot p},
\end{equation}
where $q$ is the average number of selected variables per subsample and $p$ is the number of candidates. With our threshold and large $p$, this bound remains well below one. As an additional sanity check, we repeat the entire procedure with randomly permuted labels. No neuron exceeds the selection threshold under the null distribution, confirming that the stable set reflects genuine hallucination signal rather than statistical noise.

\subsection{Causal Intervention}
\label{sec:causal-intervention}

The selection pipeline above establishes that certain neurons are reliably associated with hallucination, but association does not imply causation. To test whether the identified FH-neurons bear a causal relationship to citation hallucination, we perform activation patching at inference time.

\textbf{Intervention mechanism}
For each pro-hallucination FH-neuron $(\ell, n)$, we register a forward pre-hook on the down-projection of layer $\ell$ that intercepts the intermediate activation vector before it is projected to the residual stream. The target neuron's activation is scaled by a factor $\beta$:
\begin{equation}
\label{eq:intervention}
a_n' = \beta \cdot a_n,
\end{equation}
where $\beta = 0$ fully suppresses the neuron, $\beta = 0.5$ attenuates it, $\beta = 1$ leaves it unchanged (baseline), and $\beta > 1$ amplifies it. All other neurons remain untouched.

\textbf{Experimental conditions}
We design three conditions to establish causality from complementary directions. In the \textbf{suppression} experiment, we set $\beta < 1$ and regenerate citations across the same topics and styles used in our dataset, then re-verify each reference to measure the change in per-field hallucination rate. A decrease in hallucination rate would indicate that the suppressed neurons are causally involved in producing errors. In the \textbf{enhancement} experiment, we set $\beta > 1$ to test whether amplifying FH-neuron activations increases hallucination, providing a complementary confirmation of the causal direction. Finally, a \textbf{random control} experiment applies the same suppression ($\beta = 0$) to randomly selected neurons of the same count, repeated over five trials. A null effect in this condition would confirm that the observed changes are specific to the identified FH-neurons rather than a generic consequence of neuron ablation. All conditions use greedy decoding to ensure that any difference in output is attributable solely to the intervention. To further confirm that the intervention itself does not corrupt generation, we verified that JSON schema validity remains high across all conditions (baseline and suppression: 100\%; enhancement: 99.0\%; random control: 97.0\%), ruling out schema collapse as a confound.
% \textcolor{red}{RX: Add a generation-quality sanity check. Reviewers may ask whether strong suppression / amplification breaks fluency or the JSON output schema beyond the five verified fields. Report, for each intervention condition: (1) the fraction of outputs that parse successfully as valid JSON, and (2) optionally, perplexity or output-length change vs.\ baseline. If these already exist in the pipeline (e.g., ``valid''/``error'' counts from the older table version), simply surface them.}
\begin{table}[h]
\centering
\begin{adjustbox}{width=0.47\textwidth}
\begin{tabular}{ll c ccccc c}
\toprule
\textbf{Field} & \textbf{Direction} & \textbf{$\beta$} & \textbf{Title} & \textbf{Authors} & \textbf{Year} & \textbf{Venue} & \textbf{DOI}\\
\midrule
% ---- Title ----

\multirow{4}{*}{ Title}
& \cellcolor{lightred} Enhance $\uparrow$  & \cellcolor{lightred} 2.0  & 73.2\% & 15.5\% & 40.2\% & 42.3\% & 51.5\% \\
& \cellcolor{lightred} Enhance $\uparrow$  & \cellcolor{lightred} 4.0  & 4.7\% & 0.0\% & 10.6\% & 8.2\% & 2.4\% \\
& \cellcolor{lightgreen} Suppress $\downarrow$& \cellcolor{lightgreen} 0.0 & 82.8\% & 20.2\% & 54.5\% & 41.4\% & 55.6\%\\
& \cellcolor{lightgreen} Suppress $\downarrow$& \cellcolor{lightgreen} 0.5 & 76.5\% & 19.4\% & 48.0\% & 38.8\% & 52.0\%\\
\midrule
% ---- Authors ----
\multirow{4}{*}{Authors}
& \cellcolor{lightred} Enhance $\uparrow$  & \cellcolor{lightred} 2.0  & 77.6\% & 13.3\% & 55.1\% & 43.9\% & 59.2\%\\
& \cellcolor{lightred} Enhance $\uparrow$  & \cellcolor{lightred} 4.0  & 70.2\% & 8.5\% & 33.0\% & 24.5\% & 41.5\% \\
& \cellcolor{lightgreen} Suppress $\downarrow$& \cellcolor{lightgreen} 0.0 & 69.1\% & 23.7\% & 45.4\% & 43.3\% & 58.8\%\\
& \cellcolor{lightgreen} Suppress $\downarrow$& \cellcolor{lightgreen} 0.5 & 68.4\% & 24.2\% & 46.3\% & 43.2\% & 60.0\%\\
\midrule

% ---- Year ----
\multirow{4}{*}{Year}
& \cellcolor{lightred} Enhance $\uparrow$  & \cellcolor{lightred} 2.0   & 66.7\% & 12.5\% & 40.6\% & 32.3\% & 46.9\%\\
& \cellcolor{lightred} Enhance $\uparrow$  & \cellcolor{lightred} 4.0  & 63.8\% & 11.7\% & 27.7\% & 29.8\% & 62.8\%\\
& \cellcolor{lightgreen} Suppress  $\downarrow$ & \cellcolor{lightgreen} 0.0 & 77.8\% & 23.2\% & 46.5\% & 44.4\% & 59.6\%\\
& \cellcolor{lightgreen} Suppress  $\downarrow$ & \cellcolor{lightgreen} 0.5 & 79.4\% & 19.6\% & 44.3\% & 41.2\% & 50.5\%\\

\midrule
% ---- Venue ----
\multirow{4}{*}{Venue}
& \cellcolor{lightred} Enhance $\uparrow$ & \cellcolor{lightred} 2.0  & 78.9\% & 15.8\% & 45.3\% & 38.9\% & 49.5\%\\
& \cellcolor{lightred} Enhance $\uparrow$ & \cellcolor{lightred} 4.0  & 67.4\% & 11.6\% & 37.9\% & 31.6\% & 33.7\%\\
& \cellcolor{lightgreen} Suppress $\downarrow$ & \cellcolor{lightgreen} 0.0 & 73.2\% & 20.6\% & 40.2\% & 35.1\% & 52.6\%\\
& \cellcolor{lightgreen} Suppress $\downarrow$ & \cellcolor{lightgreen} 0.5  & 74.7\% & 20.0\% & 50.5\% & 41.1\% & 53.7\% \\
\midrule
% ---- DOI ----
\multirow{4}{*}{DOI}
& \cellcolor{lightred} Enhance $\uparrow$  & \cellcolor{lightred} 2.0   & 78.4\% & 19.6\% & 42.3\% & 42.3\% & 47.4\%\\
& \cellcolor{lightred} Enhance $\uparrow$  & \cellcolor{lightred} 4.0  & 80.4\% & 23.7\% & 57.7\% & 52.6\% & 38.1\%\\
& \cellcolor{lightgreen} Suppress $\downarrow$ & \cellcolor{lightgreen} 0.0  & 77.1\% & 18.8\% & 42.7\% & 40.6\% & 55.2\%\\
& \cellcolor{lightgreen} Suppress $\downarrow$ & \cellcolor{lightgreen} 0.5  & 77.1\% & 21.9\% & 55.2\% & 40.6\% & 57.3\%\\
\midrule
% ---- Random Control ----
\multirow{1}{*}{Random Control}
& Random  &  -  & 59.9\% & 15.3\% & 40.8\% & 35.2\% & 49.1\% \\
\multirow{1}{*}{Baseline}
& Baseline  &  -  & 76.3\% & 17.5\% & 48.5\% & 41.2\% & 54.6\% \\

\bottomrule
\end{tabular}                                                       

\end{adjustbox}
\caption{Field-level accuracy on Qwen2.5-32B-Instruct citations under neuron activation scaling. Each cell reports the percentage of citations judged correct by the verifier for that field. $\uparrow$/$\downarrow$ denotes the scaling direction (enhance/suppress). For computational feasibility, all intervention conditions and the baseline are evaluated on a fixed sample of 2 citation styles, 10 randomly selected topics, and 5 references per prompt. All values are therefore mutually comparable within this table. The baseline differs from the corresponding entry in Table~\ref{tab:hallucination_rates} due to topic-level variance in the model's parametric knowledge; see Appendix~\ref{app:topic_variation} for the full per-topic accuracy distribution.}

\label{tab:flash-verification}
\end{table}

\subsection{Results}
\label{sec:intervention-results}

Table~\ref{tab:flash-verification} reports per-field accuracy under each intervention condition; full statistical procedures are provided in Appendix~\ref{app:intervention-stats}.

The enhancement condition provides the clearest evidence for a causal link. At $\beta = 4.0$, accuracy on every targeted field drops sharply relative to baseline: Title collapses from 76.3\% to 4.7\%, Authors from 17.5\% to 8.5\%, and the remaining three fields lose between 10 and 21 percentage points. All five paired differences are negative under a Wilcoxon signed-rank test ($p = 0.031$), confirming that FH-neurons actively promote hallucination when amplified.

In the opposite direction, suppressing FH-neurons at $\beta = 0$ raises Title accuracy by 6.5 percentage points and Authors by 6.2 percentage points relative to baseline, while random ablation of the same number of neurons uniformly degrades all five fields, consistent with non-specific disruption rather than targeted correction. Comparing the two directly, FH-neuron suppression outperforms random ablation on four of five targeted fields ($p = 0.062$; narrowly above $\alpha = 0.05$ due to the small effective sample size of $n{=}5$, but consistent in direction and substantial in magnitude). We also observe partial positive spillover to non-targeted fields under moderate suppression, suggesting that bibliographic fields share some but not all of their underlying neural substrates---an asymmetry consistent with the cross-field probe transfer reported in Section~\ref{sec:probe_results}.

Not all fields respond equally to intervention. Venue-targeted suppression at $\beta{=}0$ matches the random-control level (35.1\% vs 35.2\%), yet the milder $\beta{=}0.5$ preserves accuracy at 41.1\%, suggesting that Venue's 51 FH-neurons are correctly identified but too sparsely distributed for complete ablation to remain distinguishable from random disruption. DOI enhancement at $\beta{=}4.0$ unexpectedly improves Year and Venue while degrading only DOI itself---an effect unique to DOI, whose 30 FH-neurons are the fewest of any field and concentrate in early layers (66.7\% in layers 0--21) rather than the middle-to-late layers where all other fields' neurons reside in Table~\ref{tab:fh_neuron_dist}.

\section{Related Work}
\label{sec:related}
 
\paragraph{Citation hallucination and model reliability.}
LLMs generate structurally plausible but factually incorrect references at alarming rates when operating from parametric memory alone. \cite{walters2023fabrication} found that 55\% of GPT-3.5 citations and 18\% of GPT-4 citations across 42 multidisciplinary topics were entirely fabricated, and \cite{linardon2025influence} showed that fabrication rates in GPT-4o vary sharply with topic familiarity. At the level of published literature, \cite{xu2026ghostcite} benchmarked 13 LLMs across 40 research domains and found hallucination rates ranging from 14\% to 95\%. The practical consequences are already visible: a recent audit of NeurIPS 2025 accepted papers uncovered over 100 hallucinated citations that had passed peer review undetected~\cite{gptzero2026neurips}, underscoring that this is not merely a theoretical concern but an active threat to the integrity of the scientific record. Post-hoc verification pipelines such as FactScore~\cite{min2023factscore} and SAFE~\cite{wei2024long} can identify these errors but require dozens of API calls per reference and offer no insight into why the model erred; retrieval-augmented generation~\cite{lewis2020retrieval} sidesteps the problem but is inapplicable offline. In the embodied setting, \cite{tang2026shifting} showed that standard token-level uncertainty metrics can mask safety-critical failure signals in vision-language-action models, requiring task-specific aggregation to produce reliable confidence estimates. These findings share a common theme with our work: models can appear confident while their internal signals are structured in ways that poorly reflect actual correctness, and domain-specific analysis is needed to expose and address the gap.

\paragraph{Probing and intervening on internal representations.}
A growing body of work shows that LLM hidden states encode linearly accessible truth-related signals~\cite{marks2023geometry, azaria2023internal, burns2022discovering, kossen2024semantic}, and that activation-level interventions can steer models toward truthfulness at inference time~\cite{li2023inference, chuang2023dola}. At the neuron level, \cite{gao2025hneurons} identified hallucination-associated neurons and showed that suppressing them reduces errors in factoid QA. \cite{orgad2024llm} found that truthfulness probes generalize weakly across datasets, implying that truthfulness encoding is multifaceted rather than universal. However, these studies uniformly treat hallucination as a monolithic phenomenon. Citation hallucination is harder to address: different fields within a reference can fail for different reasons. It remains unclear whether these failures share a common mechanism or are field-specific, and whether neuron-level interventions can target them individually.

\section{Conclusion}
We study citation hallucination as a structured failure of bibliographic generation, where errors decompose across fields and are internally localized. Empirically, different fields fail at different rates and follow distinct internal patterns, with signals learned for one field transferring poorly to others. Building on these findings, we identify a sparse set of field-specific hallucination neurons and show that targeted interventions on them can selectively improve citation accuracy. Overall, our results indicate that citation hallucination arises from field-dependent internal representations and can be partly mitigated from within the model, without relying solely on external retrieval.

\section{Limitations}
All intervention analyses are limited by the small number of structured fields ($n{=}5$), which constrains statistical power, and we do not evaluate whether neuron suppression affects output fluency beyond the JSON schema validity reported in Section~\ref{sec:causal-intervention}. Additionally, probing and neuron localization use a single model (Qwen2.5-32B-Instruct) and 50 topics from computer science, leaving generalization to other architectures and domains untested.

\bibliography{example_references}
\bibliographystyle{siamplain}

\appendix

\section{Effect of Citation Style on Hallucination Rate}
\label{app:style_test}

Table~\ref{tab:kruskal} reports Kruskal-Wallis test results comparing 
hallucination rates across eight citation styles for each bibliographic 
field. No field yields a significant difference ($p > 0.05$, 
$\eta^2 < 0.01$ in all cases), and the maximum observed difference 
between any two styles is less than 0.04, confirming that citation 
format has negligible practical impact on hallucination behavior.
                                
\begin{table}[h]
  \centering
  \caption{Kruskal-Wallis test results across citation styles.}
  \begin{tabular}{lrrrr}
    \toprule
    Field & H-statistic & p-value & Sig. \\
    \midrule
    Title   & 6.398 & 0.4941 & ns \\
    Authors & 5.642 & 0.5821 & ns \\
    Year    & 11.062 & 0.1359 & ns \\
    Venue   & 11.868 & 0.1050 & ns \\
    DOI     & 13.639 & 0.0580 & ns \\
    \bottomrule
  \end{tabular}
  \label{tab:kruskal}
\end{table}
\section{Serialization Example}
\label{app:serialization}

Probing hidden states at the level of individual fields requires knowing \emph{exactly} which tokens correspond to which bibliographic component. To achieve this, each generated reference is serialized into a plain-text sequence in which every field is enclosed by XML-style tags:

\begin{verbatim}
<TITLE> Attention Is All You Need </TITLE>
<AUTHORS> Ashish Vaswani |  ... </AUTHORS>
<VENUE> NeurIPS </VENUE>
<YEAR> 2017 </YEAR>
<DOI> 10.48550/arXiv.1706.03762 </DOI>
\end{verbatim}
During serialization, the character-level start and end offsets of each tagged block are recorded. For the \textsc{authors} field, individual author names (delimited by ``\texttt{|}'' separators) are further tracked as sub-spans, enabling finer-grained analysis if needed. These character offsets are later mapped to token positions via the tokenizer's offset mapping, yielding precise token spans for every field. This design ensures that downstream feature extraction operates on the exact set of tokens that encode a given field, rather than relying on heuristic or approximate boundaries.

\section{List of Computer Science Research Topics}
\label{app:topics}

The following 50 computer science research topics were used to prompt
the models for citation generation.

\begin{enumerate}
    \item Causal inference and treatment effect estimation
    \item Bayesian optimization and probabilistic modeling
    \item Reinforcement learning and policy optimization
    \item Robustness and adversarial machine learning
    \item Explainable AI and interpretability methods
    \item Representation learning and self-supervised learning
    \item Graph neural networks and graph learning
    \item Federated learning and distributed ML
    \item Fairness and bias mitigation in ML
    \item ML evaluation, benchmarks, and reproducibility
    \item Information extraction and relation extraction
    \item Question answering and retrieval-augmented generation
    \item Machine translation and multilingual NLP
    \item Summarization and factual consistency
    \item Dialogue systems and conversational agents
    \item Prompting and instruction tuning methods
    \item NLP robustness, safety, and alignment
    \item Text generation evaluation and metrics
    \item Knowledge grounding and entity linking
    \item Long-context modeling and efficient attention
    \item Image classification and representation learning
    \item Object detection and instance segmentation
    \item Vision transformers and efficient vision models
    \item 3D vision and point cloud understanding
    \item Visual question answering and vision-language models
    \item Video understanding and temporal action recognition
    \item Generative vision models and diffusion
    \item Self-supervised learning for vision
    \item Vision robustness and adversarial attacks
    \item Medical imaging and computer-aided diagnosis
    \item Distributed systems and consensus protocols
    \item Cloud computing and serverless architectures
    \item Storage systems and key-value stores
    \item Operating systems scheduling and resource management
    \item Compilers and code optimization
    \item Systems performance modeling and profiling
    \item Datacenter networking and traffic engineering
    \item GPU systems and ML systems optimization
    \item Fault tolerance and reliability engineering
    \item Observability and telemetry systems
    \item Network security and intrusion detection
    \item Cryptography and secure computation
    \item Privacy-preserving data analysis and differential privacy
    \item Malware analysis and reverse engineering
    \item Web security and authentication protocols
    \item Database query optimization and indexing
    \item Data integration and entity resolution
    \item Human-computer interaction and usability studies
    \item Program analysis and static/dynamic analysis
    \item Approximation algorithms and randomized algorithms
\end{enumerate}

\section{Output Volume Analysis}
\label{app:Output_Volume_Analysis}
\begin{figure}[h]
  \centering
  \includegraphics[width=0.45\textwidth]{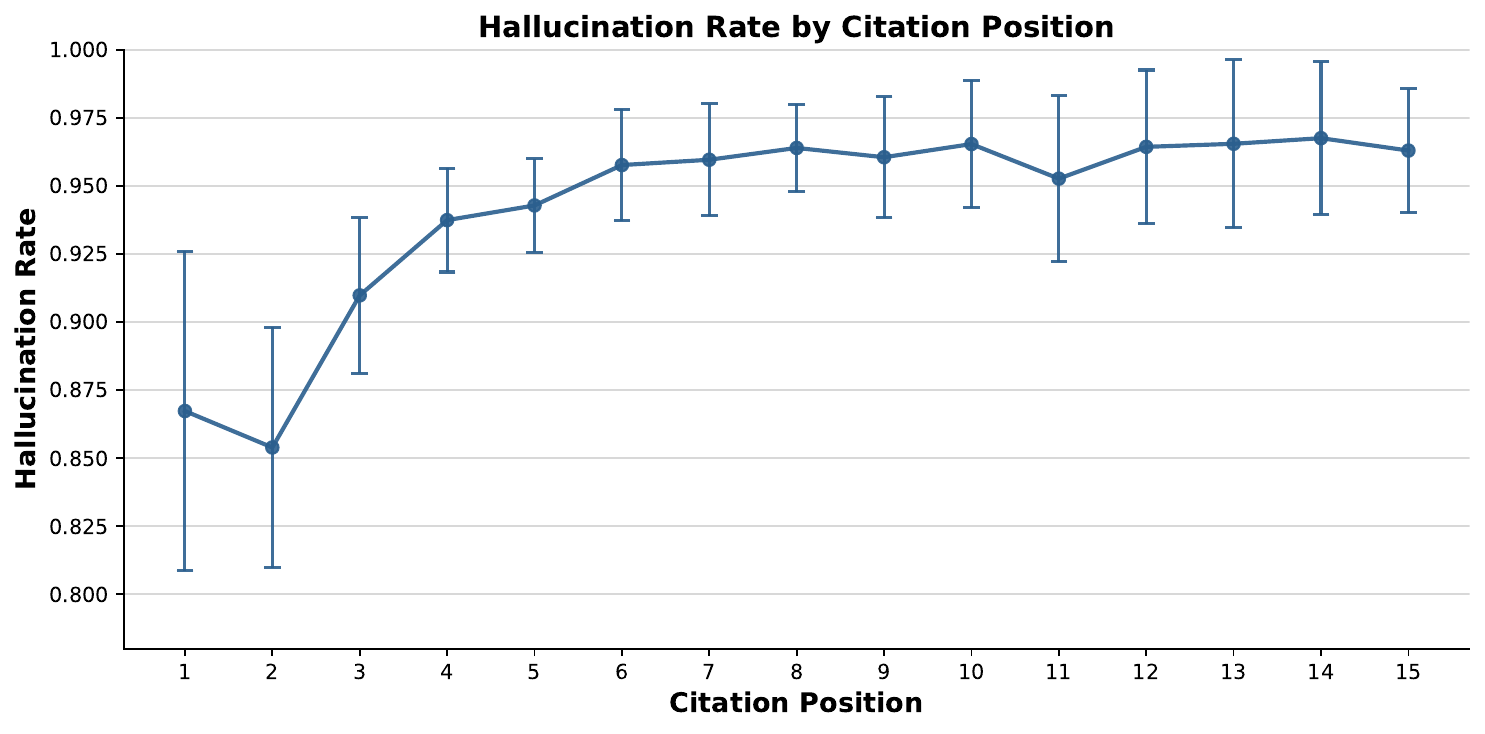}
  \caption{Hallucination rate by citation position, aggregated across all
models and generation volumes. Each bar reflects the spread of
hallucination rates observed across different experimental settings. Hallucination rate increases sharply from the second to the third citation position, then plateaus.}

  \label{fig:position}
\end{figure}

To examine whether citation order within a prompt affects reliability, we analyze hallucination rate as a function of citation position across all models and generation volumes. As shown in Figure~\ref{fig:position}, the first two citations exhibit notably lower hallucination rates, after which performance rapidly degrades and plateaus from position 3 onward. This suggests that models exhaust their most reliable parametric memory early, with subsequent references increasingly fabricated.

\section{Replication on Mistral-Small-24B}
\label{app:Mistral-Small-24B}
\begin{figure}[h]
  \centering
  \includegraphics[width=0.45\textwidth]{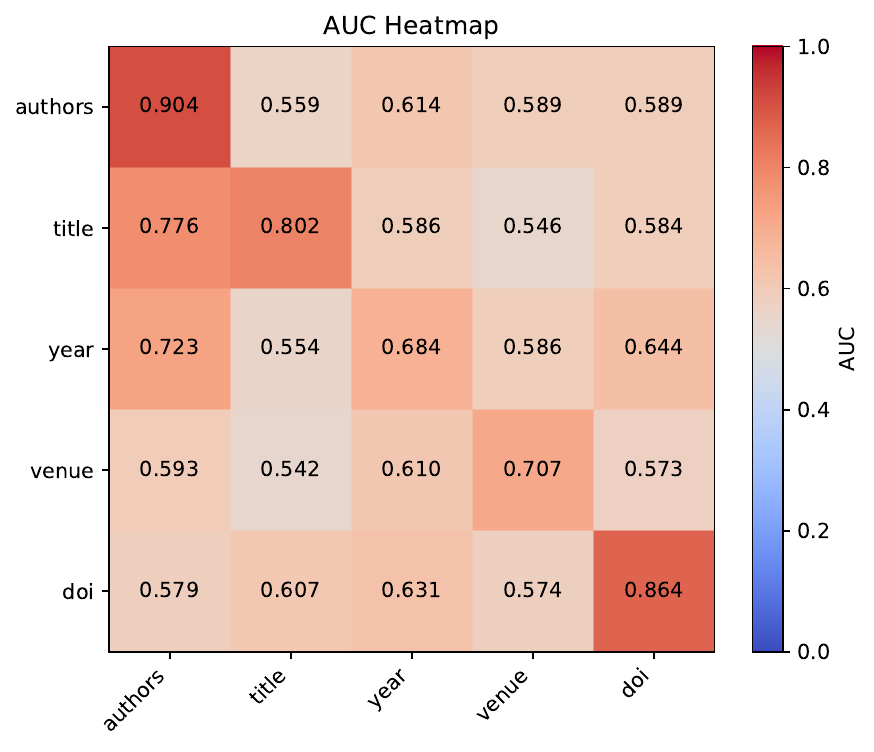}
  \caption{Cross-field AUC heatmap on Mistral-Small-24B-Instruct-2501. In-field performance remains strong, but off-diagonal transfer is higher than in Qwen2.5-32B-Instruct, indicating weaker field-specific separation.}
  \label{fig:mistral}
\end{figure}
We repeated the cross-field probe transfer analysis on Mistral-Small-24B-Instruct-2501. The resulting heatmap in Figure~\ref{fig:mistral} shows that hallucination remains linearly decodable across all five bibliographic fields, with strong in-field AUC on the diagonal. However, compared with Qwen2.5-32B-Instruct, off-diagonal transfer is noticeably higher, indicating weaker field-specific separation and greater overlap among field representations. This suggests that the existence of decodable hallucination signals generalizes across model families, while the degree of field-specific modularity is model-dependent.

\section{FH-Neuron Distribution}
\label{app:fh_neurons}

Table~\ref{tab:fh_neuron_dist} reports the number of 
positive-weight FH-neurons retained after stability 
selection for each bibliographic field, along with their 
distribution across layer terciles.

\begin{table}[h]
\centering
\caption{FH-neuron count and layer distribution per field.
Terciles: Early (layers 0--21), Mid (layers 22--42), Late (layers 43--63).
Total candidates per field: 1{,}769{,}472.}
\label{tab:fh_neuron_dist}
\setlength{\tabcolsep}{2pt}
\small
\begin{tabular}{lrrrrr}
\toprule
Field & FH-neurons & Sparsity & Early & Mid & Late \\
\midrule
Title   & 224 & 0.013\% & 24.6\% & 33.9\% & 41.5\% \\
Authors &  78 & 0.004\% & 12.8\% & 60.3\% & 26.9\% \\
Year    & 129 & 0.007\% & 16.3\% & 37.2\% & 46.5\% \\
Venue   &  51 & 0.003\% & 49.0\% & 19.6\% & 31.4\% \\
DOI     &  30 & 0.002\% & 66.7\% & 23.3\% & 10.0\% \\
\bottomrule
\end{tabular}
\end{table}

\section{Statistical Analysis of Causal Intervention}
\label{app:intervention-stats}

This appendix details the statistical procedures underlying the causal intervention results reported in Section~\ref{sec:intervention-results}. All tests treat the five bibliographic fields (Title, Authors, Year, Venue, DOI) as paired observations, with accuracy on the \emph{targeted} field, i.e.\ the diagonal entry of Table~\ref{tab:flash-verification}, as the dependent variable.
% \textcolor{red}{RX: Previously this referenced a non-existent label \texttt{tab:causal-intervention}; I corrected it to \texttt{tab:flash-verification}. Please confirm this is the intended table.} We use the one-sided Wilcoxon signed-rank test throughout; with $n = 5$ and all ranks assigned to the same sign, the minimum attainable $p$-value is $1/2^5 = 0.031$.

\paragraph{Test 1: Enhancement effect.}
We compare the targeted-field accuracy under Enhancement $\beta = 4.0$ against Baseline for each field.

\begin{center}
\setlength{\tabcolsep}{3pt}
\small
\begin{tabular}{lccc}
\toprule
\textbf{Field} & \textbf{Baseline} & \textbf{Enhance $\beta{=}4.0$} & \textbf{$\Delta$} \\
\midrule
Title   & 76.3\% & 4.7\%  & $-71.6$ \\
Authors & 17.5\% & 8.5\%  & $-9.0$  \\
Year    & 48.5\% & 27.7\% & $-20.8$ \\
Venue   & 41.2\% & 31.6\% & $-9.6$  \\
DOI     & 54.6\% & 38.1\% & $-16.5$ \\
\bottomrule
\end{tabular}
\end{center}

All five differences are negative ($H_0$: enhancement does not decrease accuracy; $H_1$: enhancement decreases accuracy). The test yields $p = 0.031$; we reject $H_0$ at $\alpha = 0.05$. The mean degradation is $-25.5$ percentage points.

\paragraph{Test 2: Random ablation effect.}
We compare Random Control accuracy against Baseline to characterize the effect of ablating an equal number of randomly selected neurons at $\beta = 0$.

\begin{center}
\begin{tabular}{lccc}
\toprule
\textbf{Field} & \textbf{Baseline} & \textbf{Random} & \textbf{$\Delta$} \\
\midrule
Title   & 76.3\% & 59.9\% & $-16.4$ \\
Authors & 17.5\% & 15.3\% & $-2.2$  \\
Year    & 48.5\% & 40.8\% & $-7.7$  \\
Venue   & 41.2\% & 35.2\% & $-6.0$  \\
DOI     & 54.6\% & 49.1\% & $-5.5$  \\
\bottomrule
\end{tabular}
\end{center}

All five differences are again negative ($p = 0.031$). Crucially, the degradation is uniform across fields with no field showing improvement, indicating non-specific disruption of generation capacity rather than targeted interference with hallucination circuitry.

\paragraph{Test 3: Specificity of FH-neuron suppression.}
We compare FH-neuron Suppression at $\beta = 0$ against Random Control on each field's targeted accuracy to determine whether FH-neuron suppression yields improvements beyond what random ablation produces.

\begin{center}
\setlength{\tabcolsep}{2pt}
\small
\begin{tabular}{lccc}

\toprule
\textbf{Field} & \textbf{FH-Suppress $\beta{=}0$} & \textbf{Random} & \textbf{$\Delta$} \\
\midrule
Title   & 82.8\% & 59.9\% & $+22.9$ \\
Authors & 23.7\% & 15.3\% & $+8.4$  \\
Year    & 46.5\% & 40.8\% & $+5.7$  \\
Venue   & 35.1\% & 35.2\% & $-0.1$  \\
DOI     & 55.2\% & 49.1\% & $+6.1$  \\
\bottomrule
\end{tabular}
\end{center}

Four of five differences are positive; Venue is essentially tied ($\Delta = -0.1$ pp). Excluding the tied pair yields an effective sample size of $n = 4$, for which the one-sided Wilcoxon signed-rank test gives $p = 0.062$. Although this narrowly exceeds the conventional $\alpha = 0.05$ threshold due to the small sample size, the effect is consistent in direction and substantial in magnitude, with a mean improvement of $+8.6$ percentage points across all five fields. When considered alongside the strong enhancement result in Test~1 and the non-specific degradation pattern of random ablation in Test~2, the combined evidence supports the conclusion that FH-neuron suppression produces field-selective improvements that are qualitatively distinct from random neuron removal.

\section{Per-Topic Accuracy Distribution}
\label{app:topic_variation}
\begin{figure*}[h]
\centering
\includegraphics[width=\textwidth]{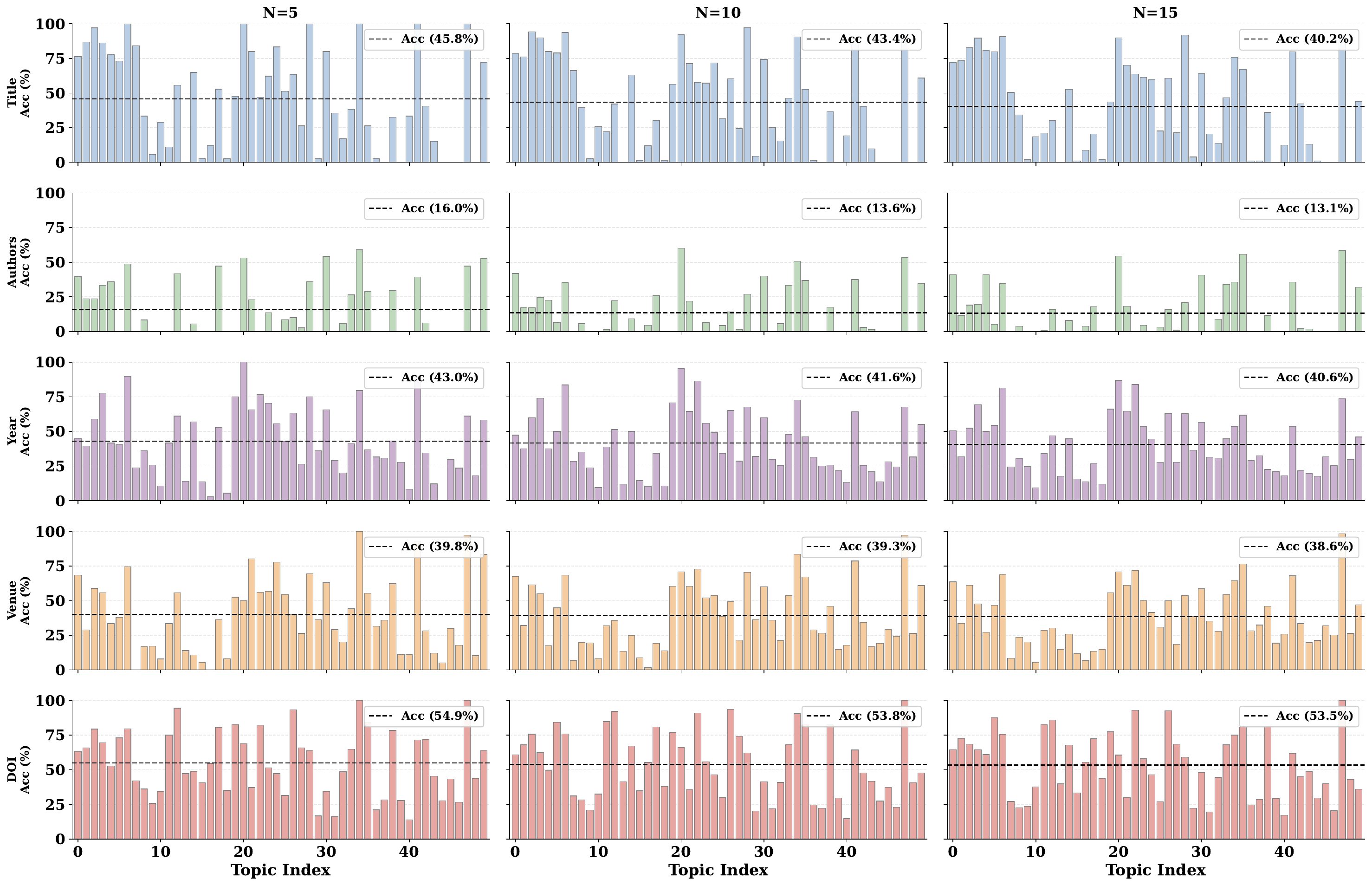}
\caption{Per-topic accuracy on Qwen2.5-32B-Instruct 
across the full 50-topic set, shown for the first $N$ 
references of each prompt ($N \in \{5, 10, 15\}$). 
Dashed lines mark the per-panel mean reported in 
Table~\ref{tab:hallucination_rates}. Accuracy is 
relatively stable across $N$ but varies substantially 
across topics, which explains why the baseline in 
Table~\ref{tab:flash-verification} (evaluated on a 
sampled topic subset) can differ from the 50-topic 
average.}
\label{fig:per_topic_acc}
\end{figure*}
To clarify the baseline discrepancy between 
Table~\ref{tab:hallucination_rates} and 
Table~\ref{tab:flash-verification}, we report per-topic 
accuracy across the full 50-topic set in 
Figure~\ref{fig:per_topic_acc}. Accuracy varies 
substantially across topics, reflecting uneven coverage 
of research areas in the model's parametric memory. 
Title accuracy in particular ranges from near 0\% on 
some topics to over 90\% on others, with comparable 
spread observed for Year, Venue, and DOI. Because our 
intervention experiments are conducted on a randomly 
sampled subset of topics for computational feasibility, 
the baseline accuracy observed in 
Table~\ref{tab:flash-verification} can differ from the 
50-topic average in 
Table~\ref{tab:hallucination_rates}. Within 
Table~\ref{tab:flash-verification} itself, however, all 
intervention conditions are evaluated on the same topic 
sample, so the reported differences across conditions 
remain internally consistent and unaffected by topic 
composition.

\section{Statistical Analysis of Layer-wise Probe Performance}
\label{app:layer_stats}

This appendix reports the statistical tests underlying the 
layer-wise probe analysis in Section~\ref{sec:layer_signal}.

\paragraph{Spearman rank correlation.} 
Table~\ref{tab:spearman} reports the Spearman correlation 
between layer index and probe AUC for each field. The null 
hypothesis is that AUC has no monotonic association with 
layer depth ($\rho = 0$).

\begin{table}[h]
\centering
\caption{Spearman correlation between layer index and 
probe AUC per bibliographic field.}
\label{tab:spearman}
\begin{tabular}{lrrr}
\toprule
Field & $\rho$ & $p$-value & Trend \\
\midrule
DOI     & $+0.763$ & $3.87 \times 10^{-7}$ & Increasing \\
Authors & $+0.600$ & $2.83 \times 10^{-4}$ & Increasing \\
Title   & $+0.554$ & $9.97 \times 10^{-4}$ & Increasing \\
Venue   & $+0.371$ & $3.66 \times 10^{-2}$ & Increasing \\
Year    & $-0.516$ & $2.52 \times 10^{-3}$ & Decreasing \\
\bottomrule
\end{tabular}
\end{table}

\paragraph{Pairwise comparison of trends.} 
Table~\ref{tab:fisher_z} reports Fisher $z$-test $p$-values 
comparing the Spearman $\rho$ of each field pair. The null 
hypothesis is $\rho_i = \rho_j$.

\begin{table}[h]
\centering
\caption{Fisher $z$-test $p$-values for pairwise comparison 
of layer--AUC trends. Significant results ($p < 0.05$) in 
bold.}
\setlength{\tabcolsep}{1pt}
\label{tab:fisher_z}
\begin{tabular}{lccccc}
\toprule
        & Title & Authors & Year & Venue & DOI \\
\midrule
Title   & ---   & 0.793   & \textbf{$<$0.001} & 0.371 & 0.150 \\
Authors &       & ---     & \textbf{$<$0.001} & 0.248 & 0.238 \\
Year    &       &         & ---                & \textbf{$<$0.001} & \textbf{$<$0.001} \\
Venue   &       &         &                    & ---   & \textbf{0.020} \\
\bottomrule
\end{tabular}
\end{table}

\paragraph{Global permutation test.} To test whether the 
five fields share the same layer--AUC relationship, we use 
the variance of the five Spearman $\rho$ values as a test 
statistic and generate a null distribution by permuting 
field labels across all (layer, AUC) pairs (10{,}000 
permutations). The observed variance (0.205) exceeds all 
permuted values ($p < 0.0001$), confirming that the 
layer-wise profiles differ across fields.

\paragraph{Bootstrap confidence intervals for peak layer.} 
Table~\ref{tab:bootstrap_peak} reports bootstrap 95\% 
confidence intervals for the peak-AUC layer per field 
(10{,}000 resamples with Gaussian noise 
$\sigma{=}0.001$ to break ties).

\begin{table}[h]
\centering
\caption{Bootstrap 95\% CI for peak layer per field.}
\label{tab:bootstrap_peak}
\begin{tabular}{lrrr}
\toprule
Field & Observed Peak & Median & 95\% CI \\
\midrule
Year    & L2  & L2  & $[2,\; 2]$   \\
Authors & L46 & L46 & $[44,\; 46]$ \\
DOI     & L48 & L48 & $[46,\; 48]$ \\
Title   & L64 & L64 & $[62,\; 64]$ \\
Venue   & L4  & L36 & $[4,\; 48]$  \\
\bottomrule
\end{tabular}
\end{table}

\section{Probe Training Details}\label{app:probe_details}
We adopt a linear probe because it tests a stronger claim than nonlinear alternatives: that hallucination status is not merely encoded in the hidden states, but is linearly separable in representation space. This follows the methodology established in prior probing work~\cite{marks2023geometry, azaria2023internal}, where linear separability is treated as evidence that a concept is represented as a direction rather than a nonlinear manifold. Within each train/test partition, examples are downsampled to a 1:1 positive-to-negative ratio to address class imbalance, ensuring that probe accuracy is not inflated by majority-class prediction. For cross-field evaluation, a probe trained on field $i$ is applied to the hidden states extracted for field $j$, and performance is measured by AUC. Concretely, a probe trained to detect author hallucination is evaluated on the hidden states of title, venue, year, and DOI spans. If cross-field AUC remains high, the fields share a common hallucination representation; if it drops to near chance (AUC $\approx$ 0.5), each field's hallucination signal occupies a distinct subspace, confirming field-specific rather than shared encoding.

\end{document}